\documentclass{article}

\usepackage[dblblindworkshop, final]{neurips_2026}
\workshoptitle{{SLMs for Agentic Systems, Paris, France, 2026}}
\usepackage[utf8]{inputenc} % allow utf-8 input
\usepackage[T1]{fontenc}    % use 8-bit T1 fonts
\usepackage{hyperref}       % hyperlinks
\usepackage{url}            % simple URL typesetting
\usepackage{booktabs}       % professional-quality tables
\usepackage{amsfonts}       % blackboard math symbols
\usepackage{nicefrac}       % compact symbols for 1/2, etc.
\usepackage{microtype}      % microtypography
\usepackage{xcolor}         % colors
\usepackage{graphicx}
\usepackage{amsmath}
\title{STEdgeFormer: MCU and NPU Deployable Transformer Architecture}

\author{%
  Sudhakar Sah \\
  STMicroelectronics, Canada\\
  \texttt{sudhakar.sah@gmail.com} \\
}

\begin{document}

\maketitle

\begin{abstract}

% Transformer-based sequence models have achieved strong results in language and time-series modeling, but self-attention remains costly for edge deployment. We propose \textit{STEdgeFormer}, a lightweight alternative of Transformers for Embedded AI accelerators that builds on polynomial mixture and DyT normalization and adds an accelerator-aware causal memory block for hardware-friendly long-sequence modeling. \textit{STEdgeFormer} achieves a perplexity of 7.39 (quantized) with 4.12M parameters and 5.34 (FP32) with 30M parameters on TinyStories, 94\% accuracy on ECG5000, and 93\% on Speech Commands using a smaller 1.39M parameters model. We further demonstrate quantized deployment of this model on the Neural Art NPU using  4.12 MB Flash and 6.51 MB RAM, although not state of the art performant, is a practical alternative to attention-based models deployable on model Neural Accelerators such as STM32N6.

Transformer-based sequence models have achieved strong performance in language and time-series modeling, but self-attention remains computationally and memory intensive for deployment on resource-constrained edge accelerators. We propose \textit{STEdgeFormer}, a lightweight, hardware-aware sequence architecture designed for efficient execution on embedded neural accelerators. \textit{STEdgeFormer} combines polynomial mixture layers and DyT normalization with an accelerator-aware causal memory block, replacing pairwise self-attention with compact causal aggregation. On TinyStories, the fully NPU-supported \textit{STEdgeFormer (MW)} achieves a perplexity of 7.39 with 4.12M parameters and 6.51 MB RAM, with an end-to-end latency of 40 ms. In comparison, the standard 4.12M-parameter configuration achieves a lower perplexity of 6.79 but requires CPU fallback for unsupported operators, resulting in 1450 ms latency. Vanilla transformer on the other hand achieves 5.71 perplexity score using 4.40M parameter model but it can't be deployed on the hardware. On downstream tasks, \textit{STEdgeFormer} achieves 94\% accuracy on ECG5000 and 93\% accuracy on Speech Commands with 1.39M parameters. These results demonstrate that hardware-aware architectural choices and operator compatibility can have a substantial impact on practical inference latency, beyond model size and task accuracy alone. Rather than targeting state-of-the-art accuracy on individual benchmarks, \textit{STEdgeFormer} is designed to provide a deployable sequence modeling architecture under the memory, computational, and operator constraints of embedded neural accelerators.

\end{abstract}

\section{Introduction}

Transformer architectures have become a widely used approach for sequence modeling because self-attention effectively captures long-range dependencies \cite{vaswani2017attention}. However, deploying Transformer models on edge devices remains challenging because self-attention scales quadratically with sequence length, while practical implementations rely on computationally and memory-intensive operations such as multi-head self-attention, softmax, masking, and large matrix multiplications that may be inefficient or unsupported on embedded NPUs \cite{vaswani2017attention}. This limitation has motivated a range of efficient sequence models, including state-space models such as S4 and Mamba \cite{gu2022efficiently,gu2024mamba}, recurrent and implicit-convolution approaches such as RWKV and Hyena \cite{peng2023rwkv,poli2023hyena}, and more recently Polynomial Mixer (PoM) and Dynamic Tanh (DyT), which provide efficient alternatives for sequence mixing and normalization, respectively \cite{picard2025pom,zhu2025dyt}.

For embedded deployment, computational complexity is only part of the challenge. Many edge accelerators support a restricted set of operators, typically favoring linear layers, convolutions, element-wise arithmetic, pooling, and standard activation functions, while operations commonly used in Transformer-style models including attention, softmax, dynamic indexing, scan/cumulative-sum operations, causal masking, and reduction-heavy normalizationmay fall back to the host CPU. Such fallbacks can increase data movement, latency, and energy consumption, making hardware-aware model design essential. To address this challenge, we propose \textit{STEdgeFormer}, a hardware-aware sequence model for embedded AI accelerators that builds on PoM and DyT and introduces an \textit{accelerator-aware causal memory block}. The proposed architecture replaces hardware-unfriendly sequence operations with accelerator-friendly primitives, enabling efficient execution on resource-constrained embedded platforms such as the STM32N6~\cite{stm32n6_ds}.

\textit{STEdgeFormer} combines compact latent-state projection, input-dependent gating, Polynomial Mixer (PoM), static multi-window causal memory aggregation, and Dynamic Tanh (DyT) normalization in an NPU-oriented architecture for edge deployment. Our contributions are as follows: (1) we introduce \textit{STEdgeFormer}, a hardware-aware sequence model that combines PoM, DyT, and static multi-window causal memory; (2) we evaluate its sequence-modeling capability on TinyStories language modeling~\cite{eldan2023tinystories}, ECG5000~\cite{dau2019archive_ucr_ecg}, and Speech Commands~\cite{warden2018speechcommands}; (3) we investigate the effect of DyT on training stability and its compatibility with embedded AI accelerators; and (4) we demonstrate quantized deployment on the Neural-ART NPU of the STM32N6~\cite{stm32n6_ds}, showing the feasibility of deploying Transformer-style sequence models on resource-constrained neural accelerators.

\section{Related Work}

\subsection{Efficient Transformers}

Transformer models use self-attention to capture long-range dependencies \cite{vaswani2017attention}, but its quadratic complexity with respect to sequence length has motivated alternatives such as linear and sparse attention, as well as compact Transformer variants for edge deployment. While these approaches reduce computational or memory costs, many still rely on attention-specific operations, including softmax, masking, and large matrix multiplications, that can be inefficient or unsupported on embedded NPUs. Consequently, reducing the asymptotic complexity of attention alone does not necessarily guarantee efficient execution on resource-constrained accelerators.

\subsection{State Space and Sequence Models}

State-space models provide an alternative approach to sequence modeling by representing historical information through latent states. S4 and Mamba~\cite{gu2022efficiently,gu2024mamba} achieve efficient long-context sequence modeling through structured state representations and selective state updates, while RWKV and Hyena~\cite{peng2023rwkv,poli2023hyena} explore recurrent and implicit-convolution formulations. Despite their favorable scaling properties, deployment on highly constrained embedded accelerators can remain challenging because of operator complexity, state-management requirements, or limited support in embedded AI frameworks.

\subsection{Polynomial and Normalization-Based Models}

Recent work has also explored alternative approaches to sequence mixing and normalization. Polynomial Mixer (PoM)~\cite{picard2025pom} replaces attention with polynomial-based sequence mixing, providing a linear-complexity alternative to self-attention, while DyT~\cite{zhu2025dyt} introduces Dynamic Tanh as an alternative to conventional normalization layers such as LayerNorm. Building on these ideas, \textit{STEdgeFormer} combines PoM and DyT with an accelerator-aware causal memory block, providing a Transformer-style sequence architecture composed primarily of primitives suitable for embedded AI accelerators.

\subsection{Edge AI Deployment}

Edge AI accelerators require models that jointly consider task performance, memory footprint, latency, and operator compatibility. Embedded NPUs typically support a restricted set of operators, making attention-heavy architectures and models containing unsupported sequence operations difficult to deploy efficiently. Rather than optimizing only the algorithmic complexity of sequence modeling, \textit{STEdgeFormer} is designed around the operator constraints of embedded NPUs by using hardware-friendly projections, element-wise nonlinearities, and causal memory aggregation. We demonstrate quantized deployment of \textit{STEdgeFormer} on the Neural-ART NPU of the STM32N6~\cite{stm32n6_ds}, highlighting its potential as a practical Transformer-style sequence model for resource-constrained edge platforms.

\section{STEdgeFormer Architecture}

\begin{figure}
\centering
\includegraphics[width=0.7\linewidth]{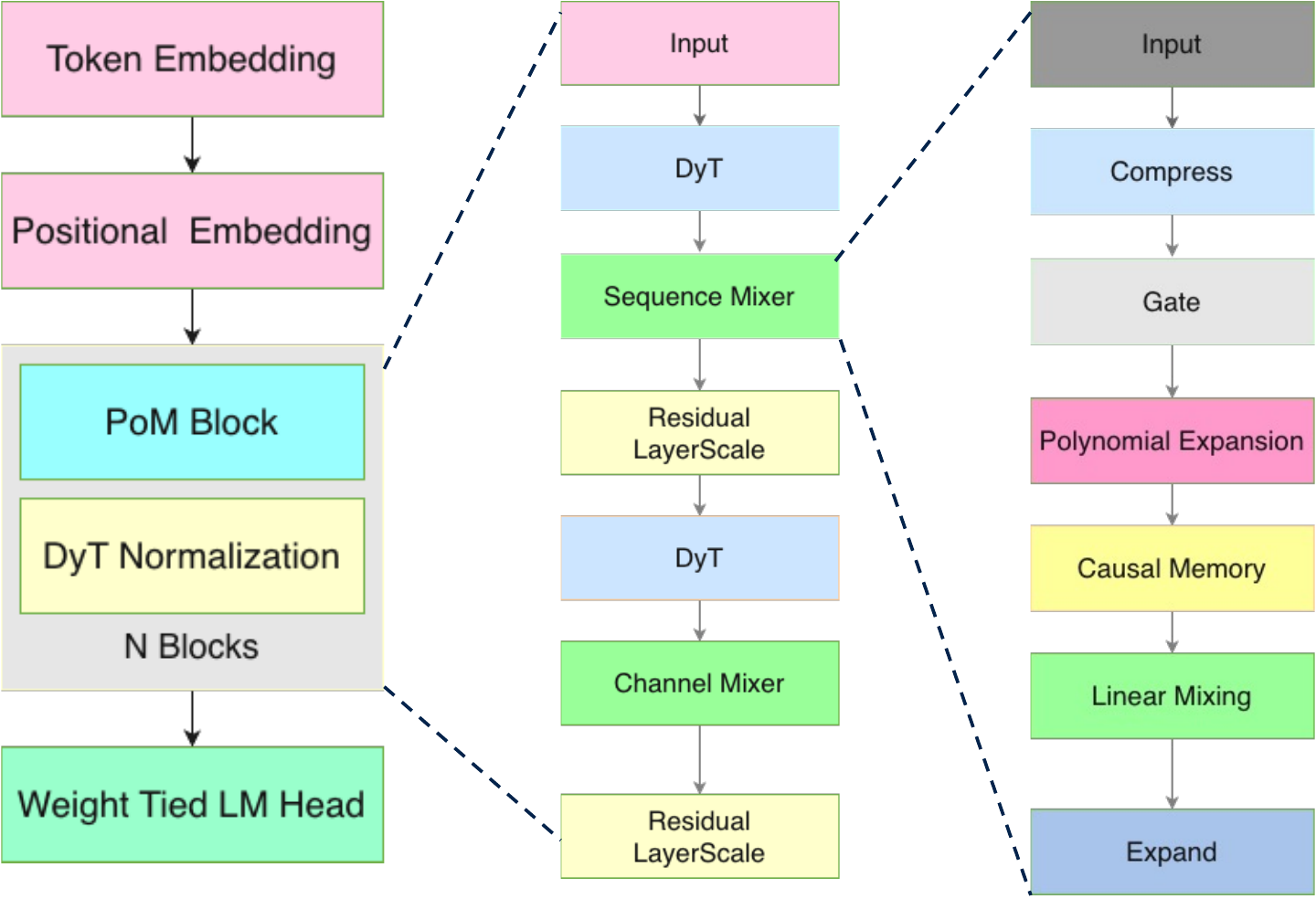}
\caption{STEdgeFormer architecture.}
\label{fig:stedgeformer_arch}
\end{figure}

As shown in Figure~\ref{fig:stedgeformer_arch}, \textit{STEdgeFormer} replaces self-attention with a compact latent state, input-dependent gating, polynomial feature mixing, and static multi-window causal memory. Given $X\in\mathbb{R}^{B\times L\times D}$, each block consists of a sequence mixer and a channel mixer with DyT and residual connections.

\subsection{Sequence Mixer}

The input is first projected to a compact latent state,
\begin{equation}
Z=XW_c,\quad W_c\in\mathbb{R}^{D\times d_s},\quad d_s=D/2,
\end{equation}
followed by residual input-dependent gating and polynomial expansion:
\begin{equation}
G=\sigma(ZW_g+b_g),\quad
\hat Z=Z+G\odot Z,\quad
P_k=\hat Z^k,; k\in{1,2,3}.
\end{equation}

\subsection{Static Multi-Window Causal Memory}

For each polynomial feature $P_k$, the causal memory operator computes moving averages over multiple temporal windows:
\begin{equation}
M_w(P_k)*t=
\frac{1}{\ell_t}
\sum*{i=\max(1,t-w+1)}^{t}(P_k)*i,\quad
\ell_t=t-\max(1,t-w+1)+1.
\end{equation}
We use four windows, $w\in{8,32,64,L}$, where $L$ represents full-history aggregation. Their outputs are combined using learnable coefficients:
\begin{equation}
S_k=\sum*{w\in{8,32,64,L}}\alpha_w M_w(P_k),
\qquad k\in{1,2,3}.
\end{equation}
The polynomial orders are then projected and combined:
\begin{equation}
Y=\sum_{k=1}^{3}a_kW_k(S_k),
\qquad W_k\in\mathbb{R}^{d_s\times D}.
\end{equation}

\subsection{Channel Mixer}

The channel mixer applies a two-layer MLP with GELU and dropout:
\begin{equation}
\mathrm{CM}(x)=
\mathrm{Dropout}!\left(W_2\mathrm{GELU}(W_1x)\right).
\end{equation}
The complete block uses pre-normalized residual updates:
\begin{equation}
x\leftarrow x+\gamma_1\mathrm{Seq}(\mathrm{DyT}(x)),\qquad
x\leftarrow x+\gamma_2\mathrm{CM}(\mathrm{DyT}(x)),
\end{equation}
where $\gamma_1$ and $\gamma_2$ are learnable layer-scale parameters.

\subsection{Complexity}

Self-attention requires $O(L^2D)$ computation and quadratic memory for token interactions. In contrast, \textit{STEdgeFormer} performs causal aggregation in the compact latent space with $O(MLd_s)$ computation, where $M=4$ is the number of aggregation windows and $d_s=D/2$. Thus, for a fixed number of windows, the proposed sequence mixer scales linearly with sequence length while avoiding explicit construction of an $L\times L$ attention matrix, making it suitable for embedded AI accelerators with limited SRAM and restricted operator support.

\section{Small Language Model Evaluation}

We evaluate \textit{STEdgeFormer} on TinyStories~\cite{eldan2023tinystories}, a synthetic language modeling benchmark designed for small language models. All experiments use a SentencePiece tokenizer with a vocabulary size of 4096. Training uses AdamW with a learning rate of $5\times10^{-4}$, weight decay of 0.01, betas $(0.9,0.95)$, gradient clipping of 1.0, batch size 128, 300 epochs, 10 warmup epochs, and a cosine \texttt{SequentialLR} schedule. Training is performed on eight NVIDIA A100 GPUs.

\subsection{Baselines and Model Variants}

We compare \textit{STEdgeFormer} against a vanilla causal Transformer baseline with 8 layers, an embedding dimension of 192, 8 attention heads, an FFN dimension of 768, GELU activation, LayerNorm, causal multi-head attention, and a sequence length of 128. In addition to the baseline, we evaluate four \textit{STEdgeFormer} variants: \textit{STEdgeFormer-S} (2.85M parameters), \textit{STEdgeFormer} (4.12M), \textit{STEdgeFormer-L} (6.55M), and \textit{STEdgeFormer-XL} (30.0M). The number of layers, embedding dimensions, and sequence lengths for these variants are reported in Table~\ref{tab:tinystories}.

\subsection{Language Modeling Results}

On TinyStories, STEdgeFormer demonstrates the trade-off between language-model quality and hardware-efficient execution. The standard STEdgeFormer achieves a PPL of 6.79 with 4.12M parameters and 6.23 MB RAM, but incurs 1450 ms latency due to CPU fallback for unsupported operators. In contrast, the fully NPU-supported STEdgeFormer (MW) achieves a PPL of 7.39 with the same model size while requiring 6.51 MB RAM and reducing latency to only 40 ms. Larger configurations can achieve lower perplexity, such as 5.34 for STEdgeFormer-XL, but rely on external memory and are therefore unsuitable for efficient on-device execution. In contrast, the vanilla Transformer cannot be deployed end-to-end on the STM32N6 because several of its layers use operators that are unsupported by the Neural-ART accelerator.These results demonstrate that hardware-aware architectural choices can substantially reduce inference latency while maintaining competitive language-model performance under embedded memory and accelerator constraints. Latency numbers for large and CPU fallback models are not reported as latency can be as high as few minutes. 

\begin{table*}[t]
\centering
\footnotesize
\setlength{\tabcolsep}{3pt}
\caption{Speech Commands v0.2 Results}
\label{tab:speechv02}
\begin{tabular}{lcc}
\toprule
Model & Accuracy (\%) & Flash / RAM (MB) \\
% Model & Accuracy (\%) & Parameters/RAM(MB) \\
\midrule
DS-CNN~\cite{zhang2017hello_dscnn} & 94.68 & 1.40/0.497 \\
Attention RNN~\cite{wen2021encoder_attentionrnn} & 96.90 & 0.202/- \\
DenseNet-BiLSTM~\cite{wen2021encoder_attentionrnn} & 97.30 & 0.223/- \\
NoisyDARTS-TC14~\cite{wen2021encoder_attentionrnn} & 97.40 & 0.107/- \\
NAS2~\cite{wen2021encoder_attentionrnn} & 97.50 & 0.886/- \\
NAO2-12~\cite{wen2021encoder_attentionrnn} & 97.50 & 0.274/- \\
NAO2-16~\cite{wen2021encoder_attentionrnn} & 97.92 & 0.458/- \\
SIDC-KWS~\cite{lim2025sidc} & 96.80 & 0.401/- \\
STEdgeFormer & 93.00 & 1.39/0.332 \\
\bottomrule
\end{tabular}
\end{table*}

\begin{table*}[t]
\centering
\caption{TinyStories Language Model Results. Parameters and RAM are reported in MB, and latency is reported in ms.}
\label{tab:tinystories}
\begin{tabular}{lcccccc}
\toprule
Model & Params/RAM & Seq. Len. & Depth & Dim. & PPL & Latency (ms) \\
\midrule

% TinyStories-SLM-GPT (HF)
% & 22.0/- & 128 & 6 & 1536 & 10.9 & -- \\

STEdgeFormer-XL$^{\oplus,\uparrow}$
& 30.0/- & 512 & 12 & 256 & 5.34 & - \\

STEdgeFormer-S$^{\oplus,\uparrow}$
& 2.85/- & 128 & 8 & 128 & 9.73 & - \\

STEdgeFormer-L$^{\oplus,\uparrow}$
& 6.55/- & 128 & 8 & 256 & 6.03 & -) \\

Transformer (Vanilla)$^{\downarrow}$
& 4.40/- & 128 & 8 & 192 & 5.71 & unsupported \\

STEdgeFormer$^{\oplus}$
& 4.12/6.23 & 128 & 8 & 196 & 6.79 & 1450 \\

\textbf{STEdgeFormer (MW)}$^{\dagger}$
& 4.12/6.51 & 128 & 8 & 196 & 7.39 & 40 \\

\bottomrule
\end{tabular}

\vspace{1mm}

\begin{minipage}{0.98\textwidth}
\footnotesize
$^{\dagger}$ Models fully supported on hardware, with all layers running on NPUs.\\
$^{\oplus}$ Models supported on hardware, but a few layers fall back to the CPU.\\
$^{\uparrow}$ Models using slower external memory and latency is very high so not reported.\\
$^{\downarrow}$ Models incompatible with unsupported layers on NPUs.
\end{minipage}

\end{table*}

\subsection{Quantized Deployment}

We benchmarked these models on the STM32N6 using the ST Developer Cloud~\cite{st_edge_ai_developer_cloud}, which provides access to the target Neural-ART accelerator and associated deployment tools. The trained and quantized models were compiled and deployed through the cloud-based STM32AI workflow, and we measured on-target memory usage and inference latency under the same deployment configuration.

For eg., we quantize the 4.12M-parameter model using static ONNX Runtime QDQ quantization. The W8/A16 configuration, in which weights are quantized to 8 bits while selected sensitive activations are retained at 16 bits, results in only a modest increase in perplexity from 7.36 to 7.39 for the MW model. This indicates that \textit{STEdgeFormer} maintains stable language modeling performance under integer quantization. 

\section{Time-Series Classification}

\subsection{ECG5000}

To assess generality beyond language modeling, we evaluate \textit{STEdgeFormer} on ECG5000, a univariate electrocardiogram classification benchmark~\cite{dau2019archive_ucr_ecg}. The classifier uses a sequence length of 128, input dimension of 1, latent dimension of 192, and six \textit{STEdgeFormer} blocks, followed by global average pooling and a lightweight prediction head. For edge deployment, the implementation avoids unsupported operators and relies primarily on linear projections and causal aggregation.

\begin{table}[t]
\centering
\caption{ECG5000 Classification Comparison.}
\label{tab:ecg}
\begin{tabular}{lcc}
\toprule
% Model & Accuracy(\%) & Memory / Size \\
Model & Accuracy (\%) & Flash / RAM \\
\midrule
E-NanoHydra \cite{cioflan2025nanohydra} & 94.13 & 3.82 kB peak memory \\
A-NanoHydra \cite{cioflan2025nanohydra} & 94.47 & 12.57 kB peak memory \\
Pruned DT (INT-8) \cite{dau2019archive_ucr_ecg} & 92.40 & Fits L1 cache ($<512$ kB) \\
TCN Baseline \cite{ecg_TCN} & 94.20 & $\sim$60 kB low RAM footprint \\
LSTM-FCN Baseline \cite{karim2018lstm_fcn} & 94.10 & $\sim$1.62 MB static RAM \\
ECG-CBA (CNN-BiLSTM) \cite{algo18110674} & 99.60 & $\sim$4.8 MB static RAM \\
STEdgeFormer & 94.00 & 1.39 MB Flash / 487 KB RAM \\
\bottomrule
\end{tabular}
\end{table}

Table~\ref{tab:ecg} presents a preliminary comparison of \textit{STEdgeFormer} with existing approaches on ECG5000 using the standard train-test split. The current \textit{STEdgeFormer} configuration achieves 94.0% accuracy with 1.39~MB of Flash, 487~kB of RAM, and 128~ms inference latency on the target platform. This accuracy is comparable to several lightweight approaches, including E-NanoHydra (94.13%), TCN (94.20%), and LSTM-FCN (94.10%), although these models have substantially different memory requirements and computational characteristics. A-NanoHydra achieves slightly higher accuracy (94.47%) with substantially lower peak memory, while ECG-CBA reports higher accuracy at the cost of a considerably larger memory footprint.

The current \textit{STEdgeFormer} implementation has not yet undergone task-specific architecture or memory optimization. Therefore, these results should be viewed as an initial baseline rather than a final optimized configuration. The objective of this experiment is not to demonstrate superior ECG classification accuracy, but to evaluate whether the proposed Transformer-style architecture can provide reasonable sequence-classification performance while remaining compatible with constrained embedded accelerators. The current results also indicate opportunities for future optimization of the Flash and RAM footprint through architecture and deployment refinements.

\subsection{Speech Commands Classification}

We also evaluate \textit{STEdgeFormer} on Speech Commands v0.2, a keyword spotting benchmark for edge speech recognition~\cite{warden2018speechcommands}. Speech is converted into sequential acoustic features and treated as a sequence classification task. The model uses the same lightweight design as in ECG5000, consisting of an input projection, a stack of \textit{STEdgeFormer} blocks, temporal aggregation, and a compact classification head. The configuration uses 12 classes, 40 input features, a sequence length of 101, an embedding dimension of 192, and a depth of 6.

Table~\ref{tab:speechv02} shows that \textit{STEdgeFormer} achieves 93.0\% accuracy on the 12-class Speech Commands benchmark. Although this accuracy is lower than that of the reference models listed in the table, the experiment serves primarily as a proof-of-concept demonstration of end-to-end execution on the target embedded platform using hardware-friendly operators. The objective is not to outperform specialized keyword-spotting architectures, but to demonstrate that the proposed Transformer-style sequence model can be applied to speech classification while satisfying the operator constraints of the embedded accelerator. These preliminary results provide a foundation for future optimization of model size, latency, memory footprint, and task-specific accuracy.

\section{Discussion}

The results across TinyStories, ECG5000, and Speech Commands demonstrate the feasibility of \textit{STEdgeFormer} as a hardware-aware alternative to attention-based sequence models. On TinyStories, \textit{STEdgeFormer} achieves reasonable language modeling performance while supporting end-to-end execution on the target embedded accelerator. In particular, the fully NPU-supported \textit{STEdgeFormer (MW)} configuration reduces latency from 1450 ms for the standard configuration, which relies on CPU fallback for unsupported operators, to 40 ms while maintaining the same model size. The ECG5000 and Speech Commands experiments further demonstrate that the same architecture can be applied to time-series and speech classification tasks. These results suggest that the combination of compact latent projection, residual gating, polynomial mixing, and causal memory provides a flexible sequence modeling architecture for resource-constrained edge platforms.

Importantly, the goal of this work is not to minimize model size, maximize task accuracy, or minimize latency in isolation, but to validate a sequence architecture that can execute end-to-end on embedded hardware under practical operator, memory, and execution-time constraints. The reported models should therefore be viewed as initial operating points that demonstrate feasibility rather than fully optimized deployments. The TinyStories results in particular illustrate the importance of hardware compatibility: avoiding CPU fallback can substantially reduce end-to-end latency even when the underlying model size remains unchanged. Future work will investigate further reductions in model footprint, latency, and memory usage through architecture tuning, quantization-aware training, and operator-level optimization.

From a computational perspective, \textit{STEdgeFormer} replaces pairwise self-attention with static multi-window causal aggregation. Standard causal self-attention requires $O(L^2d)$ computation and $O(L^2)$ storage for token interactions, whereas the proposed aggregation operates in a compact latent space with complexity $O(MLd_s)$, where $M$ is the number of fixed temporal windows and $d_s<d$ is the latent dimension. With a fixed number of windows, the proposed sequence mixer therefore scales linearly with sequence length while avoiding explicit construction of an $L\times L$ attention matrix. This design is particularly relevant to microcontroller-class NPUs, where SRAM capacity, supported operators, memory bandwidth, and operator execution efficiency jointly determine end-to-end latency.

The proposed architecture also uses a relatively small set of hardware-friendly operations, including linear projections, element-wise multiplication and addition, nonlinear activation functions, and static causal aggregation. This reduces reliance on attention-specific operations such as softmax, dynamic masking, and large token-to-token matrix multiplications, simplifying deployment on embedded accelerators with restricted operator support. More importantly, restricting the architecture to supported operations can avoid CPU fallback, which can otherwise dominate end-to-end inference latency even when the computational workload of the model itself is modest.

Compared with other efficient sequence models, \textit{STEdgeFormer} occupies a different point in the design space. State-space models such as S4 and Mamba~\cite{gu2022efficiently,gu2024mamba}, as well as recurrent and convolutional alternatives such as RWKV~\cite{peng2023rwkv} and Hyena~\cite{poli2023hyena}, provide efficient mechanisms for sequence modeling but may involve specialized state updates, scan operations, or kernels that are not directly supported by some embedded AI frameworks. In contrast, \textit{STEdgeFormer} is designed around a small set of accelerator-friendly primitives and explicitly considers operator compatibility during architectural design. This makes it a practical option for deployment scenarios in which hardware compatibility, execution simplicity, and predictable latency are important considerations.

Overall, the experiments demonstrate that \textit{STEdgeFormer} can provide reasonable performance across language modeling, physiological signal classification, and keyword spotting while supporting deployment on a constrained embedded accelerator. The TinyStories experiments further show that hardware-aware operator selection can have a substantial impact on practical inference latency: the fully NPU-supported configuration achieves 40 ms latency compared with 1450 ms when unsupported operators require CPU fallback. The current models are not optimized for minimum parameter count, memory footprint, latency, or task-specific accuracy. Nevertheless, the results establish the feasibility of the proposed hardware-aware sequence architecture and provide a basis for further joint optimization of model efficiency and accelerator execution.

\section{Conclusion}

We introduced \textit{STEdgeFormer}, a hardware-aware sequence model for embedded AI accelerators that combines polynomial mixing, DyT normalization, and an accelerator-aware causal memory block. The proposed architecture demonstrates reasonable performance across TinyStories, ECG5000, and Speech Commands while enabling end-to-end deployment on the Neural-ART NPU of the STM32N6. In particular, the quantized \textit{STEdgeFormer} model demonstrates on-device execution with measured Flash, RAM, and inference latency, highlighting the feasibility of deploying Transformer-style sequence models on resource-constrained embedded accelerators.

The primary contribution of this work is hardware-compatible sequence modeling rather than state-of-the-art task accuracy or minimum model size. The reported configurations are initial operating points and have not been extensively optimized for model size, latency, or memory footprint. Some task results remain below specialized baselines, and the current evaluation covers relatively small sequence datasets. In addition, the aggregation configuration differs across tasks, reflecting the different requirements of language, physiological signal, and speech sequence modeling.

Future work will focus on architecture tuning, quantization-aware training, operator-level optimization, and systematic evaluation on larger and more diverse sequence benchmarks. Further investigation of memory aggregation strategies and task-specific architectural adaptations may also improve the accuracy-efficiency trade-off while preserving compatibility with embedded AI accelerators.

% \input{sections/10limitations_conclusion}

%%%%%%%%%%%%%%%%%%%%%%%%%%%%%%%%%%%%%%%%%%%%%%%%%%%%%%%%%%%%

\appendix

%%%%%%%%%%%%%%%%%%%%%%%%%%%%%%%%%%%%%%%%%%%%%%%%%%%%%%%%%%%%

\newpage

\clearpage  % TODO REVIEW/FINAL: This \clearpage needs to be removed from both review and camera-ready versions.

% ---- Bibliography ----
%
% BibTeX users should specify bibliography style 'splncs04'.
% References will then be sorted and formatted in the correct style.
%
\bibliographystyle{plainnat}
\bibliography{main}

% \newpage
% \null
% \thispagestyle{empty}
% \newpage

% \input{checklist}
\end{document}